\documentclass[runningheads]{llncs}

\usepackage[year=2026]{eccv}
\usepackage[top=2cm,bottom=2cm,left=1.8cm,right=1.8cm]{geometry}
\usepackage[fontsize=12pt]{fontsize}

\usepackage{eccvabbrv}
\usepackage{graphicx}
\usepackage{booktabs}
\usepackage{amsmath,amssymb}
\usepackage{multirow}
\usepackage{xcolor}
\usepackage[accsupp]{axessibility}
\usepackage{algorithm}
\usepackage{algorithmic}
\usepackage{float}

\usepackage[pagebackref,breaklinks,colorlinks,citecolor=eccvblue]{hyperref}

\usepackage{orcidlink}

\newcommand{\ours}{IdGlow}

\renewcommand{\and}{, }
\begin{document}

\title{IdGlow: Dynamic Identity Modulation for Multi-Subject Generation}

\titlerunning{IdGlow}

\author{
Honghao Cai$^{*}$\inst{1,2},
Xiangyuan Wang$^{*}$\inst{1,3},
JingLi\inst{1},
Yunhao Bai\inst{1,3},
Tianze Zhou\inst{1,4},
Haohua Chen\inst{1},
Runqi Wang\inst{1},
Sijie Xu\inst{1},
ChanghaoQiao\inst{1},
ChaoHui\inst{1},
Yuyang Hao\inst{2},
Zezhou Cui\inst{2},
Yuyuan Yang\inst{2},
Wei Zhu\inst{1},
Yibo Chen\inst{1},
Xu Tang\inst{1},
Yao Hu\inst{1},
Zhen Li$^{\dagger}$\inst{2}
}

\authorrunning{H.~Cai, X.~Wang et al.}

\institute{
Xiaohongshu Inc. \and
The Chinese University of Hong Kong, Shenzhen \and
Peking University \and
Tsinghua University \\[0.5em]
\footnotesize $^{*}$Equal contribution.\quad $^{\dagger}$Corresponding author.
}

\maketitle
\pagestyle{plain}

\begin{figure}[H]
  \centering
  \includegraphics[width=\linewidth]{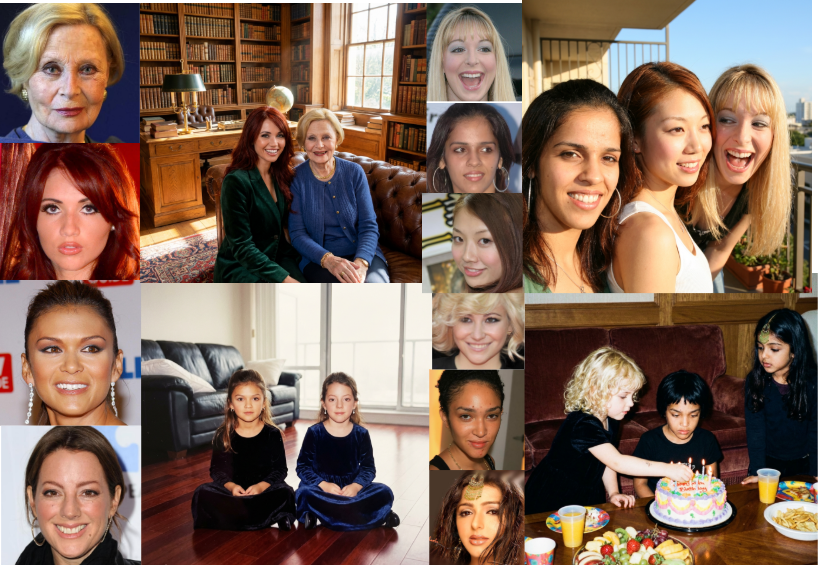}
  \caption{\textbf{Qualitative results of \ours{} on two multi-subject generation tasks.} Given a set of reference portrait images (left), \ours{} generates high-fidelity group photos (right) that faithfully preserve each individual's identity while producing coherent, aesthetically pleasing scenes. \textbf{Bottom:} Task~2---age-transformed group generation, where adult identities are transformed into child-like appearances while maintaining discriminative facial features. \textbf{Top:} Task~1---direct group fusion.}
  \label{fig:teaser}
\end{figure}

\begin{abstract}
Multi-subject image generation requires seamlessly harmonizing multiple reference identities within a coherent scene. However, existing methods relying on rigid spatial masks or localized attention often struggle with the ``stability-plasticity dilemma,'' particularly failing in tasks that require complex structural deformations, such as identity-preserving age transformation. To address this, we present \ours{}, a mask-free, progressive two-stage framework built upon Flow Matching diffusion models. In the supervised fine-tuning (SFT) stage, we introduce task-adaptive timestep scheduling aligned with diffusion generative dynamics: a linear decay schedule that progressively relaxes constraints for natural group composition, and a temporal gating mechanism that concentrates identity injection within a critical semantic window, successfully preserving adult facial semantics without overriding child-like anatomical structures. To resolve attribute leakage and semantic ambiguity without explicit layout inputs, we further integrate a badcase-driven Vision-Language Model (VLM) for precise, context-aware prompt synthesis. In the second stage, we design a Fine-Grained Group-Level Direct Preference Optimization (DPO) with a weighted margin formulation to simultaneously eliminate multi-subject artifacts, elevate texture harmony, and recalibrate identity fidelity towards real-world distributions. Extensive experiments on two challenging benchmarks---direct multi-person fusion and age-transformed group generation---demonstrate that \ours{} fundamentally mitigates the stability-plasticity conflict, achieving a superior Pareto balance between state-of-the-art facial fidelity and commercial-grade aesthetic quality.

\end{abstract}

\section{Introduction}
\label{sec:intro}

The rapid evolution of diffusion-based models~\cite{ho2020denoising,rombach2022high,song2021scorebased} has revolutionized personalized image synthesis, enabling users to generate high-quality images conditioned on specific subjects. Early personalization techniques such as DreamBooth~\cite{ruiz2023dreambooth} and Textual Inversion~\cite{gal2023an} demonstrated impressive single-subject fidelity through test-time fine-tuning. More recently, tuning-free approaches---including IP-Adapter~\cite{ye2023ipadapter}, InstantID~\cite{wang2024instantid}, PhotoMaker~\cite{li2024photomaker}, PuLID~\cite{li2024pulid}, and Arc2Face~\cite{papantoniou2024arc2face}---have achieved zero-shot identity-preserving generation by injecting face recognition embeddings (\eg, ArcFace~\cite{deng2019arcface}) into the diffusion pipeline, eliminating the need for per-subject optimization.

While these methods excel at single-subject personalization, extending them to \emph{multi-subject} scenarios exposes a fundamental limitation in current diffusion architectures: the severe conflict between macro-structural layout and micro-identity fidelity. Multi-subject generation---particularly group photo fusion---requires simultaneously preserving the distinct facial features of every individual while composing them into a coherent, interacting scene. Existing approaches attempt to solve this by explicitly partitioning the cross-attention maps (\eg, via spatial bounding boxes or localized resamplers) to prevent identity blending. Such region-based approaches function as \emph{compositing pipelines} rather than true generative models, enforcing rigid spatial isolation that inherently precludes natural subject interaction and holistic structural transformations such as age progression. However, these static, spatial-centric constraints force the model to inject identity features uniformly across all denoising timesteps. This fundamentally ignores the internal spectral dynamics of diffusion models, where global structures and local textures are formed at distinctly different stages. Consequently, enforcing rigid identity constraints during the early layout-formation phase inevitably disrupts the structural naturalness (\eg, failing to generate child-like proportions in age transformation~\cite{or2020lifespan,alaluf2021only}), while uniform injection in late stages often leads to ``plastic'' or ``micro-adult'' artifacts. We term this the \textbf{Stability-Plasticity Dilemma}.

Meanwhile, text prompt quality has emerged as a critical yet often overlooked factor in generation quality. Recent work on prompt optimization---such as Promptist~\cite{hao2023optimizing} and Dynamic Prompt Optimizing~\cite{mo2024dynamic}---has demonstrated that LLM-rewritten prompts can substantially boost aesthetic scores and text-image alignment. However, these methods focus on general-purpose image generation and do not address the semantic precision and attribute disentanglement required for multi-subject scenes, where ambiguous prompts easily induce attribute leakage and identity confusion.

To address these challenges, we propose \textbf{\ours{}} (Dynamic Identity Modulation for Multi-Subject Generation), a progressive two-stage framework designed to harmonize identity fidelity with aesthetic quality. The core philosophy of \ours{} lies in the \emph{dynamic modulation} of identity constraints based on the generative mechanics of the diffusion process, rather than applying them as static signals.

Our contributions are three-fold:
\begin{itemize}
    \item We propose \ours{}, a progressive two-stage framework for dynamic identity modulation in multi-subject generation. This unified framework addresses the challenges of multi-subject synthesis across a wide spectrum of tasks, ranging from direct group fusion to complex structural transformations exemplified by age-transformed generation.
    \item We establish a Dynamics-Aware Identity Modulation Strategy grounded in the internal generative dynamics of the diffusion process. Unlike conventional static constraints, this strategy dictates that identity information should be injected according to the spectral evolution of denoising timesteps. Specifically, we introduce Task-Adaptive Loss Annealing to coordinate multi-subject interactions and Temporal-Gated ID Injection to resolve structural conflicts in age transformation. The latter selectively activates ID constraints only during the critical semantic window of $t \in [0.3, 0.6]$, effectively mitigating the Stability-Plasticity Dilemma by aligning loss intensity with the model's structural formation.
    \item We design a Fine-Grained Direct Preference Optimization (DPO) stage for joint identity-aesthetic alignment. By utilizing authentic multi-person group photos as positive anchors and curating preference pairs that target identity drift, multi-subject artifacts, and photorealistic degradation, this stage simultaneously recalibrates identity fidelity towards real-world distributions and ensures that all subjects are harmoniously integrated with high-fidelity, consistent textures.
\end{itemize}

\begin{figure*}[t]
  \centering
  \includegraphics[width=\linewidth]{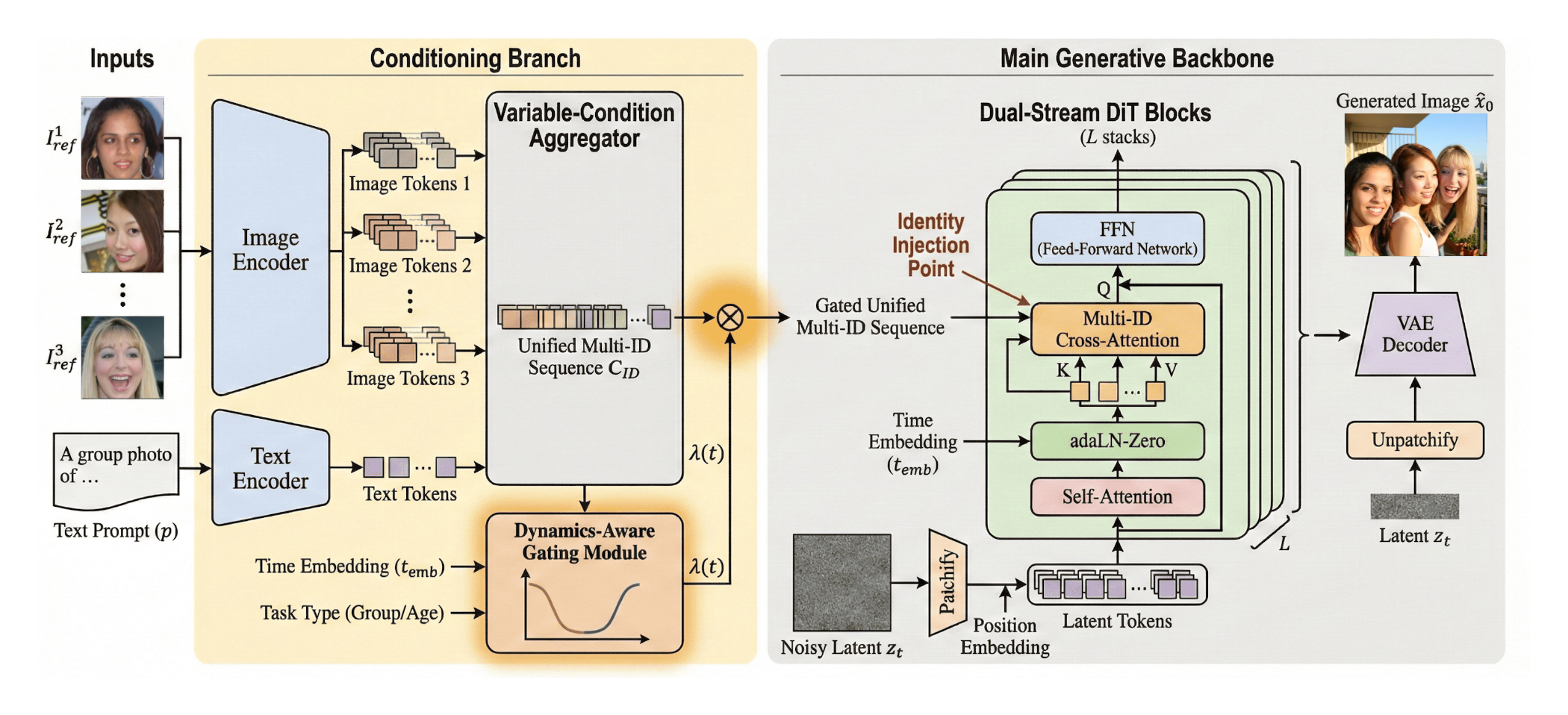}
  \caption{\textbf{The architecture of \ours{}-DiT.} The model processes variable numbers of reference identities through a unified encoding strategy, forming a concatenated multi-ID sequence. A key innovation is the Dynamics-Aware Gating Module (highlighted in orange), which modulates the intensity of the identity sequence based on the diffusion timestep $t$ and the specific task (\eg, age transformation curves). These gated features are injected into the main generation process via Dual-Stream DiT Blocks, where they serve as keys/values in a specialized Multi-ID Cross-Attention layer, interacting with the noisy latent queries.}
  \label{fig:architecture}
\end{figure*}

\section{Related Work}
\label{sec:related}

\noindent\textbf{Personalized image generation.}
The field of personalized image synthesis has seen rapid development, primarily driven by techniques such as DreamBooth~\cite{ruiz2023dreambooth} and Textual Inversion~\cite{gal2023an}. While these methods excel at single-subject personalization, extending them to multi-subject scenarios introduces significant complexities, notably identity leakage and structural blending. Recent efforts~\cite{xiao2023fastcomposer,kumari2023multi,wang2024msdiffusion,zhou2024storydiffusion} have attempted to address these issues through multi-concept fine-tuning, explicit layout-based guidance, or localized cross-attention mechanisms, while other tuning-free adapters~\cite{ye2023ipadapter,wang2024instantid,li2024photomaker,li2024pulid} primarily focus on single-subject scenarios. However, these approaches often rely on static spatial constraints that fail to account for the evolving denoising process, leading to the Stability-Plasticity Dilemma. Unlike previous works that treat identity preservation as a constant requirement, \ours{} introduces a dynamics-aware modulation strategy to balance fidelity with contextual naturalness.

\noindent\textbf{Diffusion dynamics and ID consistency.}
A growing body of literature~\cite{ho2020denoising,rombach2022high,song2021scorebased,esser2024scaling} has begun to explore the internal mechanics of the diffusion process, revealing that different timesteps contribute disproportionately to various image attributes. For instance, global structures are typically established in high-noise stages, while fine textures emerge in later stages. In the context of identity consistency, some methods have proposed timestep-dependent conditioning. However, most existing strategies employ heuristic schedules that are not specifically tailored to the structural conflicts inherent in tasks such as age transformation~\cite{or2020lifespan,alaluf2021only}. Our work, \ours{}, builds upon these insights by establishing a Temporal-Gated Injection mechanism, which selectively activates identity constraints during the critical semantic window to decouple structural priors from facial features.

\noindent\textbf{Preference optimization in AIGC.}
Direct Preference Optimization (DPO)~\cite{rafailov2023direct} has emerged as a powerful alternative to traditional reinforcement learning from human feedback~\cite{ouyang2022training}, demonstrating remarkable efficacy in aligning large language models and, more recently, text-to-image diffusion models~\cite{wallace2024diffusion,black2024training,li2024flowgrpo}. Beyond generation, preference optimization has also been applied to image editing, with works such as FireRed-Image-Edit~\cite{team2026firered}. Current applications of DPO in image generation primarily focus on improving general aesthetic scores or following complex prompts. The extension of DPO to group-level identity refinement remains largely unexplored. \ours{} fills this gap by introducing Fine-Grained Group-Level DPO, which utilizes curated preference pairs to specifically target multi-subject artifacts and texture degradation, thereby ensuring both identity fidelity and photorealistic harmony in complex group scenes.

\section{Methodology}
\label{sec:method}

\subsection{System Architecture: Dual-Stream Diffusion Transformer}
\label{sec:arch}

\ours{} adopts a dual-stream architecture based on the Diffusion Transformer (DiT)~\cite{peebles2023scalable,esser2024scaling}, designed to achieve deep fusion of textual semantics and visual features (\cref{fig:architecture}).

\subsubsection{Latent Space Representation.}
To improve computational efficiency and capture high-frequency textures, we map the original image $\mathbf{x}$ into a latent space $\mathcal{Z}$ defined by a pretrained Variational Autoencoder (VAE)~\cite{kingma2014auto}. Given an image $\mathbf{x}$, its latent representation is $\mathbf{z} = \mathcal{E}(\mathbf{x})$, where $\mathcal{E}$ denotes the VAE encoder. The generation process operates within the latent space by learning a velocity field $\mathbf{v}_\theta(\mathbf{z}_t, t, \mathbf{c})$~\cite{lipman2023flow,liu2023flow}, and the final output is decoded back to pixel space via the decoder $\mathcal{D}$.

\subsubsection{Dual-Stream Denoising Network.}
The model architecture follows a dual-stream processing paradigm. The first stream is the \emph{visual stream}, which processes latent variable sequences formed through patchification. The second stream is the \emph{semantic stream}, which takes high-level semantic embeddings $\mathbf{c}$ extracted by a Vision-Language Model (VLM) as conditional input. These two streams are deeply coupled through cross-attention mechanisms within the DiT Transformer blocks, ensuring that the generated image content remains precisely aligned with complex textual instructions.

\subsection{Task-Specific Prompt Synthesis via Badcase-Driven Alignment}
\label{sec:prompt}

In multi-subject generation, precisely controlling complex spatial layouts and subject interactions poses a significant challenge. Traditional static prompts often fail to provide sufficient structural constraints, leaving diffusion models vulnerable to the ``hallucination trap'' when handling complex contextual transformations (\eg, group fusion after age transformation). Specifically, ambiguous semantic inputs frequently induce attribute leakage (e.g., blending clothing colors or facial features) and lighting incongruity, severely degrading the aesthetic quality and identity fidelity.

To eliminate attribute ambiguity at its source, we introduce a specially optimized Image-Edit-Prompt model (denoted $\mathcal{M}_{P}$), serving as a ``semantic enricher'' bridging high-level user intent and low-level visual generation. Given a task instruction $\mathcal{I}$ and reference subjects, this model automatically synthesizes highly descriptive prompts $p = \mathcal{M}_{P}(\mathcal{I})$ enriched with detailed appearance attributes, lighting conditions, and natural interaction semantics. By grounding these rich textual details prior to the denoising process, $\mathcal{M}_{P}$ substantially alleviates the attribute disentanglement burden on the downstream DiT backbone, enabling it to focus entirely on maintaining identity fidelity and refining pixel-level textures.

However, existing Vision-Language Models (VLMs) inevitably exhibit attribute confusion when generating complex instructions. To ensure $\mathcal{M}_{P}$ provides absolutely precise semantic enrichment without cascading ambiguities to the generative pipeline, we discard the conventional instruction learning paradigm in favor of a badcase-driven preference alignment strategy.

Specifically, we establish a taxonomy for semantic ambiguities, encompassing attribute leakage, lighting inconsistency, and detail omission. Based on this, we construct a large-scale set of human preference pairs: negative samples comprise barren instructions that easily induce identity blending originally generated by the base model, while positive samples are strictly constrained, highly descriptive instructions with accurate attribute bindings. During training, by guiding the model to maximize the relative probability of richly attributed descriptions over ambiguous ones, we explicitly align the model's outputs with fine-grained visual semantics, thereby endowing $\mathcal{M}_{P}$ with robust attribute-aware instruction synthesis capabilities for complex multi-subject scenes.

As shown in \cref{tab:comparison_task1,tab:comparison_task2}, dynamic prompts driven by $\mathcal{M}_{P}$ achieve consistent gains in both FaceSim and aesthetic scores over static templates.

\begin{figure}[tb]
  \centering
  \includegraphics[width=\linewidth]{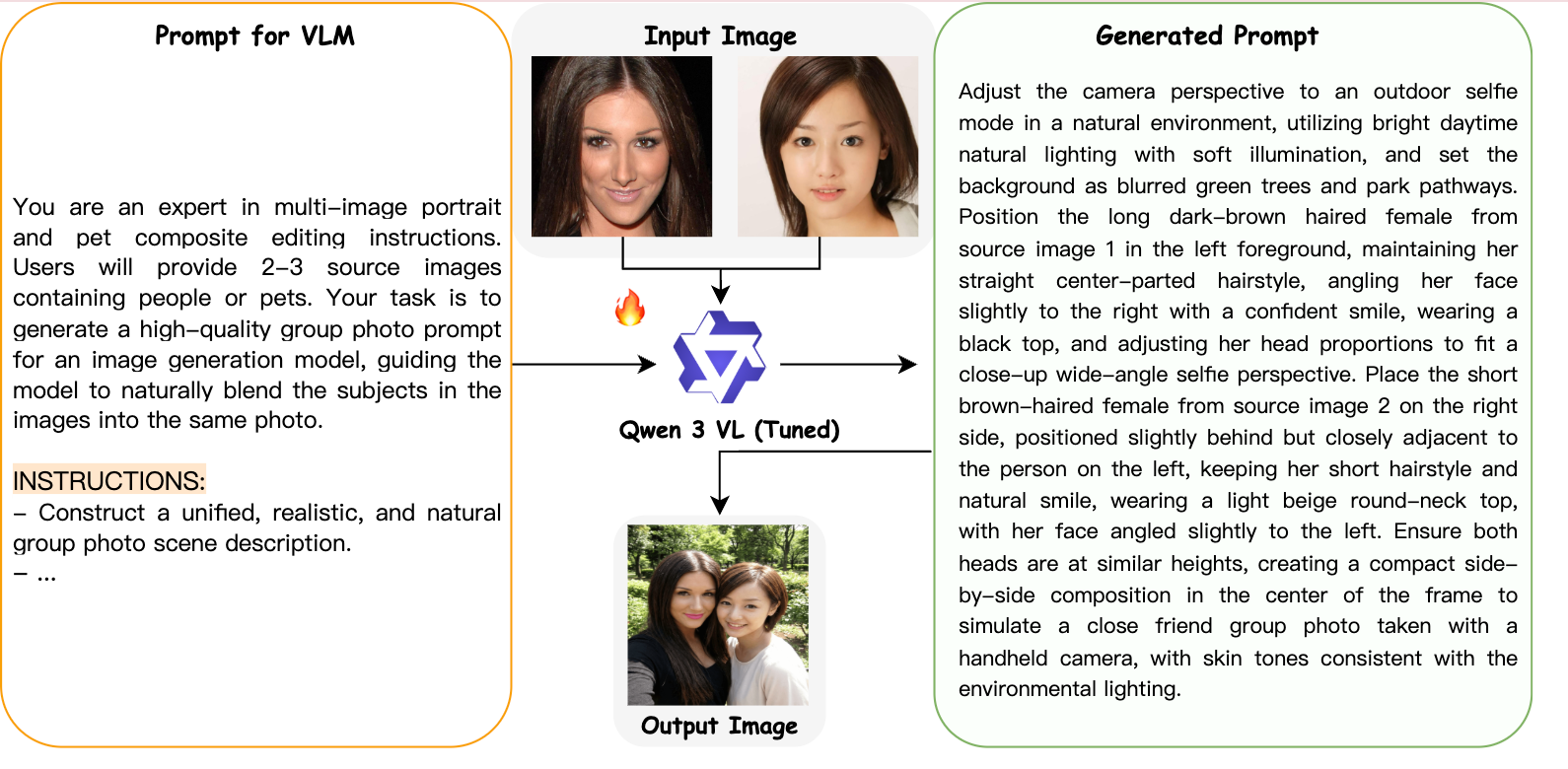}
  \caption{\textbf{Task-specific prompt synthesis via the Image-Edit-Prompt model.} Given a set of input images and a structured VLM prompt with spatial instructions, our fine-tuned Qwen~3~VL model automatically generates a detailed, spatially precise prompt that specifies subject positions, appearance attributes, and scene composition. This generated prompt then guides the diffusion model to produce a coherent group photo with correct spatial arrangement and identity preservation.}
  \label{fig:prompt_model}
\end{figure}

\subsection{Dynamics-Aware Identity Modulation Strategy}
\label{sec:id_loss}

\begin{figure*}[tb]
  \centering
  \includegraphics[width=\linewidth]{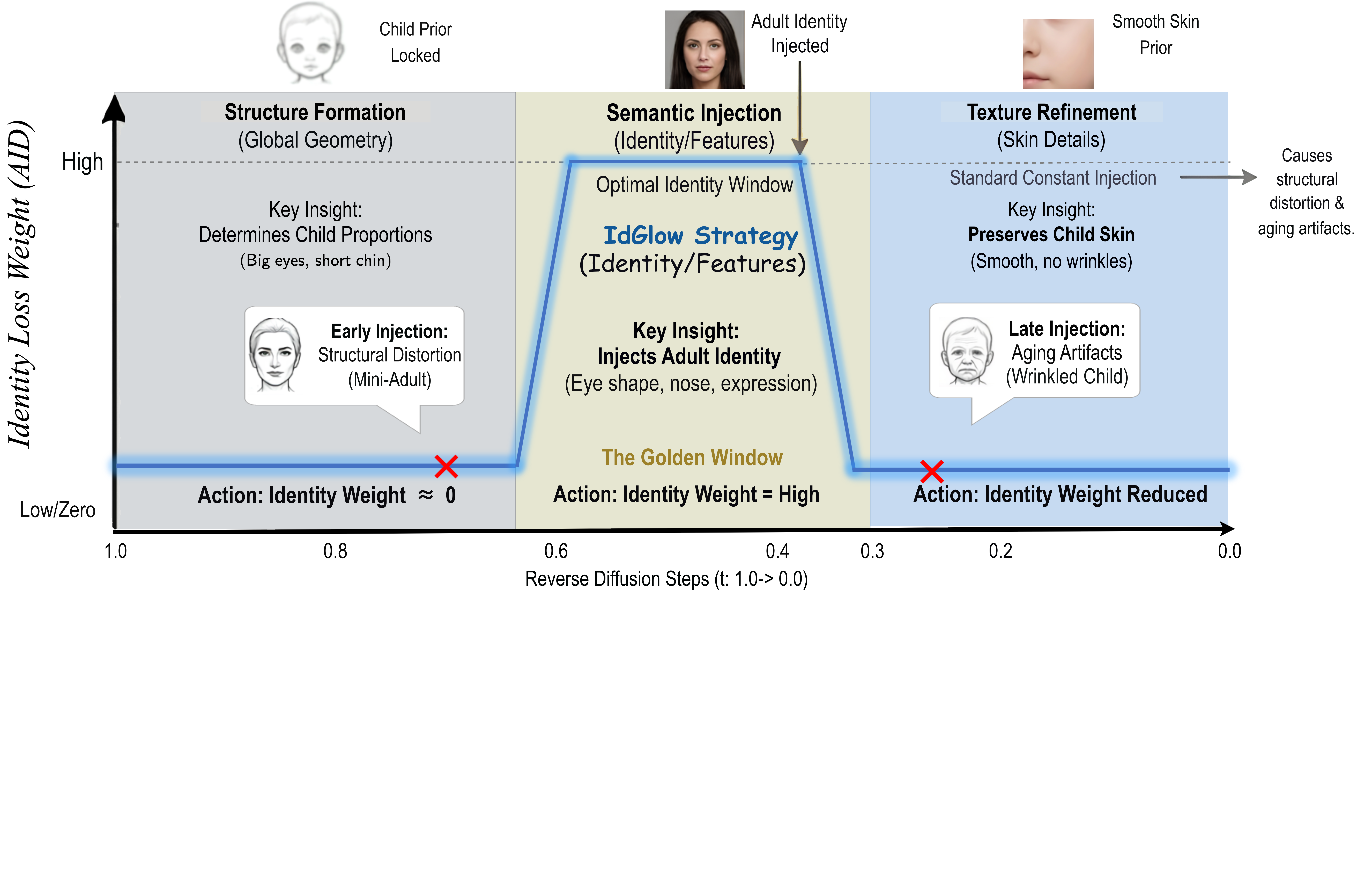}
  \caption{\textbf{Dynamics-aware identity modulation tailored to specific generative tasks.} We design dynamic loss schedules aligned with the spectral evolution of the diffusion process ($t$: $1.0 \to 0.0$). (a)~\emph{Loss Annealing for Group Generation}: high identity weight in early stages to establish identity foundation, gradually relaxed for harmonious lighting and pose. (b)~\emph{Temporal Gating for Age Transformation}: identity constraints selectively activated only during the semantic window ($t \in [0.3, 0.6]$), suppressed at early stages for child-like structure and reduced at late stages for smooth skin texture.}
  \label{fig:task_adaptive}
\end{figure*}

This constitutes the methodological core of our framework. We propose that the total loss function should be dynamically modulated according to the generative dynamics of the diffusion process. The objective function for the Supervised Fine-Tuning (SFT) stage is defined as:
\begin{equation}
    \mathcal{L}_{\text{SFT}}(\theta) = \mathbb{E}_{\mathbf{z}_0, \mathbf{c}, t} \left[ \| \mathbf{v}_\theta(\mathbf{z}_t, t, \mathbf{c}) - (\mathbf{z}_1 - \mathbf{z}_0) \|^2 + \Omega(t) \cdot \lambda_{\text{base}} \, \mathcal{L}_{\text{ID}} \right],
    \label{eq:sft_total}
\end{equation}
where $\Omega(t)$ is the temporal modulation operator that regulates identity constraint intensity according to the denoising stage, and $\mathcal{L}_{\text{ID}}$ denotes the identity loss computed via a trained face recognition encoder.

\subsubsection{Multi-Face Identity Loss via Hungarian Matching.}
In multi-subject scenarios, the generated image $\hat{\mathbf{x}}$ contains multiple faces whose spatial arrangement may differ from the input order. We therefore design a matching-based identity loss to ensure each source identity is compared against its most corresponding generated face. Specifically, given $k$ source identity images $\mathcal{I} = \{I_1, \ldots, I_k\}$, we first apply a face detector to $\hat{\mathbf{x}}$ and each $I_j$ to localize facial regions, yielding detected face crops $\{\hat{f}_1, \ldots, \hat{f}_m\}$ from the generated image and $\{f_1, \ldots, f_k\}$ from the source images. We then extract identity embeddings using a trained face recognition encoder $\mathcal{F}$ and construct a pairwise cosine similarity matrix $\mathbf{S} \in \mathbb{R}^{k \times m}$, where $S_{j,i} = \cos(\mathcal{F}(f_j), \mathcal{F}(\hat{f}_i))$. The optimal bipartite assignment is obtained via the Hungarian algorithm~\cite{kuhn1955hungarian}:
\begin{equation}
    \sigma^* = \arg\max_{\sigma} \sum_{j=1}^{k} S_{j,\sigma(j)}.
    \label{eq:hungarian}
\end{equation}
The identity loss is then computed as the average cosine distance over the matched pairs:
\begin{equation}
    \mathcal{L}_{\text{ID}} = \frac{1}{k} \sum_{j=1}^{k} \left(1 - \cos\!\big(\mathcal{F}(f_j),\; \mathcal{F}(\hat{f}_{\sigma^*(j)})\big)\right).
    \label{eq:id_loss}
\end{equation}
This formulation is essential for multi-subject generation, as it decouples identity comparison from spatial position, enabling the model to freely arrange subjects in the scene without being penalized for layout variations.

\subsubsection{Mechanism 1: Task-Adaptive Loss Annealing.}
\label{sec:linear_decay}
For direct group fusion tasks, we define a linear annealing operator:
\begin{equation}
    \Omega_{\text{anneal}}(t) = \max\!\left(0,\; \frac{t - \tau}{1 - \tau}\right),
    \label{eq:linear_decay}
\end{equation}
where $\tau$ is a threshold parameter. This enables the model to focus on establishing identity features during the initial denoising stages, while gradually releasing degrees of freedom in later stages to optimize fine-grained texture quality processed by the VAE decoder.

\subsubsection{Mechanism 2: Temporal-Gated ID Injection.}
\label{sec:temporal_gating}
For tasks involving structural shifts (\eg, adult-to-child age transformation), we design a gating operator $\Omega_{\text{gate}}(t)$ that activates only within the critical semantic window $t \in [0.3, 0.6]$:
\begin{equation}
    \Omega_{\text{gate}}(t) = \begin{cases}
        0               & \text{if } t > 0.6 \quad \text{(structure formation)} \\[2pt]
        1               & \text{if } 0.3 \leq t \leq 0.6 \quad \text{(semantic window)} \\[2pt]
        0               & \text{if } t < 0.3 \quad \text{(texture refinement)}
    \end{cases},
    \label{eq:temporal_gating}
\end{equation}
When $t > 0.6$, child-like anatomical priors form freely; when $t < 0.3$, fine textures refine without identity interference. The window $t \in [0.3, 0.6]$ is where discriminative features (eye shape, nose contour) are resolved, enabling targeted identity transfer onto the established child structure.

The complete SFT training procedure with dynamic identity loss modulation is summarized in Algorithm~\ref{alg:sft}.

\begin{algorithm}[h]
\caption{SFT Training with Dynamic Identity Loss Modulation}
\label{alg:sft}
\begin{algorithmic}[1]
\REQUIRE Pretrained DiT model $\mathbf{v}_\theta$, face encoder $\mathcal{F}$, training set $\mathcal{D}_{\text{SFT}}$, task type $\mathcal{T} \in \{\text{group}, \text{age}\}$, base weight $\lambda_{\text{base}}$, threshold $\tau$
\ENSURE Fine-tuned model $\mathbf{v}_\theta$
\FOR{each iteration}
    \STATE Sample $(\mathbf{z}_0, \mathbf{c}, \mathcal{I}) \sim \mathcal{D}_{\text{SFT}}$
    \STATE Sample timestep $t \sim \mathcal{U}(0, 1)$
    \STATE Construct noisy latent: $\mathbf{z}_t = (1 - t)\,\mathbf{z}_0 + t\,\boldsymbol{\epsilon}$, \quad $\boldsymbol{\epsilon} \sim \mathcal{N}(\mathbf{0}, \mathbf{I})$
    \STATE Predict velocity: $\hat{\mathbf{v}} = \mathbf{v}_\theta(\mathbf{z}_t, t, \mathbf{c})$
    \STATE $\mathcal{L}_{\text{flow}} = \| \hat{\mathbf{v}} - (\mathbf{z}_1 - \mathbf{z}_0) \|^2$
    \STATE \textbf{Task-adaptive modulation operator}
    \IF{$\mathcal{T} = \text{group}$}
        \STATE $\Omega(t) = \max\!\big(0,\; \frac{t - \tau}{1 - \tau}\big)$
    \ELSIF{$\mathcal{T} = \text{age}$}
        \STATE $\Omega(t) = \mathbf{1}[0.3 \leq t \leq 0.6]$ \hfill
    \ENDIF
    \STATE Decode estimate: $\hat{\mathbf{x}} = \mathcal{D}(\hat{\mathbf{z}}_0)$ via one-step prediction
    \STATE Detect faces in $\hat{\mathbf{x}}$ and $\mathcal{I}$; compute $\mathcal{L}_{\text{ID}}$ via Hungarian matching (Eq.~\ref{eq:id_loss})
    \STATE $\mathcal{L}_{\text{SFT}} = \mathcal{L}_{\text{flow}} + \Omega(t) \cdot \lambda_{\text{base}} \cdot \mathcal{L}_{\text{ID}}$
    \STATE Update $\theta$ via $\nabla_\theta \mathcal{L}_{\text{SFT}}$
\ENDFOR
\end{algorithmic}
\end{algorithm}

\subsection{Fine-Grained Group-Level DPO}
\label{sec:dpo}

In the second stage, we further improve model performance through Direct Preference Optimization (DPO)~\cite{rafailov2023direct,wallace2024diffusion}. We leverage identity similarity $S_{\text{ID}}$ provided by a face recognition model and aesthetic scores $S_{\text{AE}}$ from LAION-Aesthetics~\cite{schuhmann2022laion} to construct preference pairs.

\subsubsection{Flow-Matching Log-Likelihood Surrogate.}
For a flow-matching DiT, exact log-likelihoods are intractable. Following Diffusion-DPO~\cite{wallace2024diffusion}, we approximate the log-likelihood ratio via the MSE of the velocity prediction as a tractable surrogate:
\begin{equation}
    \log \frac{\pi_\theta(\mathbf{z}|\mathbf{c})}{\pi_{\text{ref}}(\mathbf{z}|\mathbf{c})} \approx -\frac{1}{2}\,\Delta(\mathbf{z};\theta), \quad
    \Delta(\mathbf{z};\theta) = \|\mathbf{v}_\theta - \mathbf{v}^*\|^2 - \|\mathbf{v}_{\text{ref}} - \mathbf{v}^*\|^2,
    \label{eq:mse_surrogate}
\end{equation}
where $\mathbf{v}^* = \mathbf{z}_1 - \mathbf{z}_0$ is the ground-truth flow target and $\|\cdot\|^2$ denotes the mean squared error averaged over spatial dimensions.

\subsubsection{Weighted-Margin DPO Objective.}
Standard Diffusion-DPO treats the chosen and rejected gradient contributions symmetrically. In multi-subject generation, however, improving identity fidelity on the chosen sample and suppressing artifacts on the rejected sample are tasks of asymmetric difficulty. Drawing inspiration from the asymmetric DPO strategy recently adopted in FireRed-Image-Edit~\cite{team2026firered} for image editing, We introduce a \textbf{weighted margin} formulation:
\begin{equation}
    \mathcal{L}_{\text{DPO}}(\theta) = -\mathbb{E}\!\left[\log \sigma\!\left(-\tfrac{\beta}{2}\!\left(\alpha\,\Delta(\mathbf{z}_w;\theta) - \Delta(\mathbf{z}_l;\theta)\right) + m\right)\right]
    + \lambda_{\text{SFT}}\,\mathbb{E}\!\left[\|\mathbf{v}_\theta(\mathbf{z}_{w,t},t,\mathbf{c}) - \mathbf{v}^*\|^2\right],
    \label{eq:dpo_loss}
\end{equation}
where $\alpha > 0$ is the asymmetric weight that amplifies the chosen pair's gradient contribution relative to the rejected pair, $m \geq 0$ is an additive margin enforcing a minimum quality gap between preferred and rejected outputs, and $\lambda_{\text{SFT}} \geq 0$ is an SFT regularization coefficient that anchors the policy to the chosen distribution and prevents reward hacking. Setting $\alpha=1$, $m=0$, and $\lambda_{\text{SFT}}=0$ recovers standard Diffusion-DPO~\cite{wallace2024diffusion}. This stage ensures that \ours{} transcends mere pixel-level matching, achieving commercial-grade aesthetic quality while actively refining (rather than merely maintaining) identity consistency across all subjects in the scene.

\subsubsection{Data Sources.}
\label{sec:dpo_data}

We design a multi-stage pipeline for constructing high-quality preference data to ensure effective DPO training.

\noindent\textbf{Real Data.} We collect two types of real images: high-quality single-person portraits serving as conditional inputs, and authentic multi-person group photos as positive sample references. Crucially, these authentic group photos serve as absolute upper bounds for both aesthetic harmony and identity fidelity. By contrasting them against synthetic negative samples, the DPO policy learns to correct the subtle identity shifts and micro-artifacts that typically accumulate during the SFT stage.

\noindent\textbf{Synthetic Data.} To augment the training data, we generate group photos using our model. These synthetic samples naturally vary in quality, where high-quality outputs serve as positive samples while those with degraded identity preservation or image artifacts serve as negative samples. We further apply controlled perturbations such as noise injection and saturation adjustment to real images or high-quality generated results, artificially enlarging the quality gap to construct preference pairs with clearer distinctions for more effective DPO learning.

\subsubsection{Multi-stage Data Filtering.}
\label{sec:dpo_filter}

Let $\mathcal{D}_{\text{raw}}$ denote the raw candidate preference pair set, where each sample is a quadruple $(x_w, x_l, \mathcal{I}, p)$: $x_w$ is the chosen (preferred) sample, $x_l$ is the rejected sample, $\mathcal{I} = \{I_1, I_2, \ldots, I_k\}$ is the set of input condition images where $k \in \{2, 3\}$, and $p$ is the text prompt.

We define a set of consistency metrics $\Phi = \{\phi_1, \phi_2, \ldots, \phi_n\}$ to evaluate the alignment between generated images and input conditions. These metrics include: \textbf{Identity Consistency} $\phi_{\text{id}}$ (measures facial identity preservation using embedding similarity), \textbf{Object Consistency} $\phi_{\text{obj}}$ (measures attribute preservation for objects such as clothing and accessories), \textbf{Semantic Consistency} $\phi_{\text{sem}}$ (measures global semantic alignment between input and output), and \textbf{Aesthetic Quality} $\phi_{\text{aes}}$ (measures overall image quality and visual appeal).

For each metric $\phi_i$, we compute the difference between positive and negative samples:
\begin{equation}
    \Delta_i = \phi_i(x_w, \mathcal{I}) - \phi_i(x_l, \mathcal{I}),
\end{equation}

\noindent\textbf{Stage 1: Automatic Filtering.} Using Vision-Language Models (VLM) and task-specific metrics, we filter out low-quality samples and retain only pairs with significant quality differences across one or more consistency dimensions.

\noindent\textbf{Stage 2: Human Verification.} We conduct manual review to ensure the consistency and quality of positive samples, removing pairs with obvious defects or ambiguous preference relationships.

The final filtered dataset is defined as:
\begin{equation}
    \mathcal{D} = \{ (x_w, x_l, \mathcal{I}, p) \in \mathcal{D}_{\text{raw}} \mid \exists i, \Delta_i > \tau_i \},
\end{equation}
where $\tau_i$ is the threshold for the $i$-th consistency metric.

\section{Experiments}
\label{sec:exp}

\subsection{Experimental Setup}
\label{sec:setup}

\subsubsection{Datasets and Benchmarks.}
To evaluate the effectiveness of \ours{}, we utilize the CelebA-HQ~\cite{karras2018progressive} dataset as our primary data source. Our test set is constructed to simulate diverse real-world multi-subject scenarios, covering both two-subject and three-subject configurations and totaling 250 unique high-resolution portrait identities across a range of ages, ethnicities, and genders. To ensure a controlled and fair comparison, Task~1 (Direct Group Fusion) and Task~2 (Age-Transformed Group Generation) utilize the exact same set of subject identities, allowing us to isolate the contribution of our dynamic modulation strategy on structural transformations.

\subsubsection{Evaluation Metrics.}
We employ two primary quantitative metrics to assess identity fidelity and aesthetic quality.

\textbf{FaceSim} (Identity Similarity): We measure identity similarity using the ArcFace~\cite{deng2019arcface} model, specifically the \texttt{buffalo\_l} variant from the InsightFace library. We calculate the cosine similarity between the facial features of the reference portraits and the synthesized individuals in the group scenes.

\textbf{Aesthetic Score}: To evaluate the photorealistic quality and visual appeal of the generated images, we utilize the LAION-Aesthetics~\cite{schuhmann2022laion} predictor. This model is based on CLIP ViT-L/14~\cite{radford2021learning} features and provides a standardized score reflecting the aesthetic excellence of the results.

We compare with five methods: FastComposer~\cite{xiao2023fastcomposer}, nano banana pro, Qwen-Image-Edit-2511~\cite{wu2025qwenimage} (our base model), HunyuanImage~\cite{cao2025hunyuanimage}, and Seedream~\cite{gao2025seedream}. For fair evaluation, all baseline models, including FastComposer, are executed using their official pre-trained weights and recommended default inference configurations without heuristic post-processing.

\noindent\textbf{Note on Single-Subject Adapters.} We exclude region-based single-subject methods (\eg, InstantID~\cite{wang2024instantid} with spatial masking), as they are architecturally incapable of structural age transformation in Task~2 and require manual mask engineering for Task~1, making comparison semantically invalid.

\subsection{Quantitative Comparison}
\label{sec:main_results}

We compare \ours{} against several state-of-the-art personalized generation baselines, including standard Diffusion Transformers and existing multi-subject fusion methods. We evaluate all methods under two prompting strategies: \emph{static prompts}, which use fixed templates, and \emph{dynamic prompts}, which are automatically generated by our Image-Edit-Prompt model to produce more detailed, context-aware instructions.

As summarized in \cref{tab:comparison_task1,tab:comparison_task2}, \ours{} achieves the highest FaceSim across all settings while maintaining competitive Aesthetic scores. In Task~2, our temporal-gated mechanism enables precise identity injection without compromising child-like structural integrity. The Fine-Grained DPO stage further elevates both metrics, confirming that aligning the generative manifold with real-world group photos recovers high-frequency discriminative features. \ours{} establishes a new Pareto frontier, harmonizing identity fidelity and aesthetic quality where prior methods fail.

\begin{table}[tb]
  \caption{\textbf{Quantitative comparison on Task~1 (Direct Group Fusion)} under static and dynamic prompts. FS = FaceSim, Aes = Aesthetic Score. Best results in \textbf{bold}, second best \underline{underlined}.}
  \label{tab:comparison_task1}
  \centering
  \small
  \setlength{\tabcolsep}{4pt}
  \begin{tabular}{l cccc}
    \toprule
    \multirow{2}{*}{Method}
      & \multicolumn{2}{c}{Static} & \multicolumn{2}{c}{Dynamic} \\
    \cmidrule(lr){2-3} \cmidrule(lr){4-5}
      & FS & Aes & FS & Aes \\
    \midrule
    FastComposer~\cite{xiao2023fastcomposer} & 0.16 & 2.34 & 0.18 & 2.94 \\
    Nano banana pro         & \underline{0.66} & \textbf{6.48} & \underline{0.70} & 6.21 \\
    Qwen-Image-Edit-2511~\cite{wu2025qwenimage}    & 0.52 & 4.69 & 0.59 & 5.02 \\
    HunyuanImage~\cite{cao2025hunyuanimage}            & 0.48 & 5.22 & 0.52 & \underline{6.35} \\
    Seedream~\cite{gao2025seedream}                & 0.57 & 5.71 & 0.60 & 5.95 \\
    \midrule
    \ours{} (Ours)          & \textbf{0.72} & \underline{6.21} & \textbf{0.75} & \textbf{6.48} \\
    \bottomrule
  \end{tabular}
\end{table}

\begin{table}[tb]
  \caption{\textbf{Quantitative comparison on Task~2 (Age-Transformed Generation)} under static and dynamic prompts. FS = FaceSim, Aes = Aesthetic Score. Best results in \textbf{bold}, second best \underline{underlined}.}
  \label{tab:comparison_task2}
  \centering
  \small
  \setlength{\tabcolsep}{4pt}
  \begin{tabular}{l cccc}
    \toprule
    \multirow{2}{*}{Method}
      & \multicolumn{2}{c}{Static} & \multicolumn{2}{c}{Dynamic} \\
    \cmidrule(lr){2-3} \cmidrule(lr){4-5}
      & FS & Aes & FS & Aes \\
    \midrule
    FastComposer~\cite{xiao2023fastcomposer} & 0.13 & 2.35 & 0.14 & 2.81 \\
    Nano banana pro         & \underline{0.32} & \underline{6.51} & \underline{0.32} & \textbf{6.59} \\
    Qwen-Image-Edit-2511~\cite{wu2025qwenimage}    & 0.23 & 5.21 & 0.24 & 5.48 \\
    HunyuanImage~\cite{cao2025hunyuanimage}            & 0.26 & 6.01 & 0.28 & 6.26 \\
    Seedream~\cite{gao2025seedream}                & 0.28 & 6.35 & 0.29 & \underline{6.53} \\
    \midrule
    \ours{} (Ours)          & \textbf{0.35} & \textbf{6.52} & \textbf{0.37} & 6.52 \\
    \bottomrule
  \end{tabular}
\end{table}

FastComposer's low scores stem from its SD1.5-based localized cross-attention, which imposes rigid spatial constraints that fail under close subject interactions and collapse entirely in Task~2, where child-like anatomical structures cannot be recovered from source spatial priors, yielding severe ``micro-adult'' artifacts.

\subsection{Qualitative Analysis}
\label{sec:qualitative}

Visual comparisons further underscore the superiority of our approach. Baseline methods frequently produce ``rigid'' results in direct group fusion, where faces appear incongruent with the environment's lighting. In age-transformation tasks, these methods often result in ``micro-adult'' artifacts, where adult facial features are forcefully superimposed onto small-scale face structures. In contrast, \ours{} produces harmoniously integrated group scenes with natural lighting and youthful textures. The synthesized children effectively capture the ``spirit'' of the original subjects while adhering to correct anatomical proportions, demonstrating the efficacy of our decoupled structural-semantic injection.

\subsection{Ablation Study}
\label{sec:ablation}

We conduct an ablation study to isolate the contributions of our core components (\cref{tab:ablation}).
\subsubsection{Effect of Dynamic Loss Weighting.}
Removing the loss annealing in Task~1 leads to an increase in visual artifacts and rigid expressions, as the model lacks the flexibility to adapt to complex poses.
\subsubsection{Impact of Temporal Gating Window.}
To validate our choice of the semantic window $t \in [0.3, 0.6]$, we compare three non-overlapping injection intervals on Task~2 (\cref{tab:ablation}). Injecting identity constraints during the early structural stage ($t \in [0.6, 0.9]$) severely disrupts child-like facial proportions and bone structure, producing ``doll-face adult'' artifacts that degrade both FaceSim and Aesthetic scores. Injecting only during the late texture stage ($t \in [0.1, 0.3]$) preserves child anatomy but fails to transfer geometric identity features (nose shape, eye shape), resulting in low FaceSim; meanwhile, adult skin textures (pores, roughness) leak into the output, moderately reducing Aesthetic quality. Our chosen window $t \in [0.3, 0.6]$ targets the mid-semantic stage where discriminative facial geometry forms, achieving the best balance between identity fidelity and visual quality. This confirms that identity injection must be precisely synchronized with the diffusion process's spectral dynamics.
\subsubsection{Contribution of DPO.}
While SFT establishes the structural baseline for multi-subject composition, the significant leap in FaceSim (\cref{tab:ablation}) demonstrates that our DPO stage is not merely an aesthetic filter, but a critical identity-refinement mechanism. This dual improvement is fundamentally driven by our preference data construction strategy (\cref{sec:dpo_data}). By utilizing authentic multi-person group photos as absolute positive anchors and ID-degraded synthetic outputs as negative samples, the DPO objective explicitly penalizes identity drift. Consequently, the model is forced to align its generated feature manifold with the real-world photographic distribution, simultaneously eliminating high-frequency artifacts (the ``Glow'' effect) and pulling the generated faces closer to the ground-truth identity embeddings.

\begin{table}[tb]
  \caption{\textbf{Ablation study: Effect of dynamic identity loss modulation and DPO.} ``Constant'' uses fixed $\Omega(t)=1.0$. ``Task-Adaptive'' denotes loss annealing for Task~1 and temporal gating for Task~2.}
  \label{tab:ablation}
  \centering
  \begin{tabular}{@{}llcc@{}}
    \toprule
    Task & Configuration & FaceSim  & Aesthetic  \\
    \midrule
    \multirow{4}{*}{Task 1} & SFT w/o ID Loss          & 0.45 & \textbf{6.25} \\

                            & SFT w/ Constant ID Loss   & 0.59 & 5.83 \\
                            & SFT w/ Loss Annealing     & 0.64 & 6.09 \\
                            & + Fine-Grained DPO         & \textbf{0.72} & 6.21 \\
    \midrule
    \multirow{6}{*}{Task 2} & SFT w/o ID Loss           & 0.12 & \textbf{6.71} \\
                            & SFT w/ Constant ID Loss    & 0.18 & 6.22 \\
                            & SFT w/ Gating $t \in [0.6, 0.9]$ & 0.15 & 5.89 \\
                            & SFT w/ Gating $t \in [0.1, 0.3]$ & 0.16 & 6.38 \\
                            & SFT w/ Gating $t \in [0.3, 0.6]$ & 0.28 & 6.45 \\
                            & + Fine-Grained DPO          & \textbf{0.35} & 6.52 \\
    \bottomrule
  \end{tabular}
\end{table}

\section{Conclusion}
\label{sec:conclusion}

In this paper, we presented \ours{}, a progressive framework for identity-consistent multi-subject generation. By establishing a Dynamics-Aware Identity Modulation Strategy, we demonstrated that identity constraints must be synchronized with the generative dynamics of the diffusion process. Through the integration of task-adaptive SFT for structural grounding and fine-grained group-level DPO for joint identity-aesthetic refinement, \ours{} successfully resolves the Stability-Plasticity Dilemma, delivering both high-fidelity identity preservation and superior aesthetic quality.

\bibliographystyle{splncs04}
\bibliography{main}
\end{document}